\documentclass[sigconf,nonacm]{acmart}
\AtBeginDocument{%
  }
\setcopyright{cc}
\setcctype{by}
\acmDOI{10.1145/3843282.3844436}
\acmYear{2026}
\copyrightyear{2026}
\acmISBN{979-8-4007-2985-0/2026/10}
\acmConference[AgenticDev '26]{Proceedings of the 1st International Workshop on Agentic AI for Next-Generation Software Development}{October 12--16, 2026}{Munich, Germany}
\acmBooktitle{Proceedings of the 1st International Workshop on Agentic AI for Next-Generation Software Development (AgenticDev '26), October 12--16, 2026, Munich, Germany}
\acmSubmissionID{asews26agenticdevmain-p100-p}
\received{2026-07-14}
\received[accepted]{2026-08-20}
\usepackage{pifont}
\usepackage{tcolorbox}
\tcbuselibrary{breakable,skins}
\newtcolorbox{discussionbox}[1]{
    enhanced jigsaw,
    breakable,
    colback=gray!5,
    colframe=black,
    boxrule=0.5pt,
    arc=2mm,
    left=4pt, right=4pt, top=4pt, bottom=4pt,
    before skip=8pt, after skip=8pt,
    before upper={\sloppy},
    title=\centering{#1},
    fonttitle=\bfseries
}
\begin{document}

\title{Value-Preserving Architectures for Agentic AI Systems}


\author{Alessandro Pesare}
\authornote{Both authors contributed equally to the paper.}
\orcid{0009-0003-4875-2068}
\affiliation{%
  \institution{TU Wien}
  \city{Vienna}
  \country{Austria}
}
\email{alessandro.pesare@tuwien.ac.at}

\author{Tommaso Dolci}
\authornotemark[1]
\orcid{0000-0002-1403-7766}
\affiliation{%
  \institution{TU Wien}
  \city{Vienna}
  \country{Austria}
}
\email{tommaso.dolci@tuwien.ac.at}

\author{Katja Hose}
\orcid{0000-0001-7025-8099}
\affiliation{%
  \institution{TU Wien}
  \city{Vienna}
  \country{Austria}
}
\email{katja.hose@tuwien.ac.at}

\author{Emanuel Sallinger}
\orcid{0000-0001-7441-129X}
\affiliation{%
  \institution{TU Wien}
  \city{Vienna}
  \country{Austria}
}
\email{emanuel.sallinger@tuwien.ac.at}

\begin{abstract}
The emergence of agentic AI and LLM-based multi-agent systems (MAS) presents unprecedented opportunities for automating complex tasks, while simultaneously raising critical concerns about the preservation of fundamental human-centered values, such as privacy, fairness, and safety.
Although software engineering has traditionally focused on functional correctness, the adoption of LLMs and AI agents into complex socio-technical systems has intensified the need for responsible software engineering and robust value alignment.
In MAS, architectural design decisions, such as coordination mechanisms, communication protocols, and system topologies, play a central role in shaping system behavior and the outcomes they produce.
This paper argues that architectural choices influence not only the functionality and performance of MAS but can also promote value-oriented system behavior.
Therefore, we investigate how different architectural designs support different human-centered values, discussing the following value-preserving architectural patterns: (i)~a privacy-aware architecture with a federated topology, (ii)~a distributed architecture to promote pluralism and diversity, and (iii)~a guard-agent architecture to detect and mitigate unfairness.
Finally, we introduce representative use cases to illustrate the proposed architectures in real-world scenarios.
By linking architectural design with human-centered values, this work lays the foundation for a unified set of architectural patterns and guidelines towards the design of trustworthy MAS.
\end{abstract}



\bibliographystyle{ACM-Reference-Format}

\maketitle

\section{Introduction}
Agentic AI is rapidly reshaping how software is conceived, built, and operated.
AI agents re-frame the use of large language models (LLMs) from passive text generation to specialized autonomous entities capable of planning, using external tool, and adapting according to feedback at runtime~\cite{wei2026agentica}. 
For instance, a travel-planning AI agent can autonomously search online, take actions such as booking flights and hotels, and adapt its plan when circumstances change, such as a flight cancellation.
The same principles of autonomy, planning, and tool use can be extended beyond a single agent.
Multi-agent systems (MAS) orchestrate multiple agents that collaborate through structured communication to achieve more complex objectives by distributing a common goal across several specialized agents, leveraging problem decomposition and parallel execution that exceed single agent capability~\cite{tran2025multiagent,adimulam2026orchestration}.
In particular, agentic AI and MAS have attracted significant attention in software engineering, where coordinated agents are employed for automating requirements elicitation, code synthesis, test generation, and decision support across the software lifecycle~\cite{hong2023metagpt,he2025llmbaseda}.

However, as the number of autonomous agents and the depth of their interactions increase, MAS gain more capability at the cost of reduced human oversight, diminished transparency, and weaker guarantees about the resulting software artifacts~\cite{chan2023harms}.
In the absence of both human oversight and adequate system-level design guarantees, a single failure at a coordination step can silently propagate across the entire system~\cite{greshake2023not}.
For instance, when personal data traverses agent boundaries to be processed outside their original domain, the decision-making process can violate the principle of data minimization~\cite{li2025123}.
In architectures governed by a central orchestrator agent, the system’s implicit preferences and decision policies can influence the final outcome by filtering out minority or dissenting perspectives~\cite{ashkinaze2025plurals,feng2024modular}.
At the same time, when agents are conditioned by historically biased data, a number of social biases can be amplified and propagated through successive stages of processing, ultimately producing outputs that embed discrimination while appearing fluent and authoritative~\cite{nguyen2025social,yu2025netsafe}.

While software engineering has traditionally focused on functional correctness and technical reliability, the emergence of complex AI-based socio-technical systems demands increasing attention towards ethical awareness, safety, and alignment to values such as privacy, fairness, and security~\cite{stahl2024ethicsa,lu2022roadmap}.
In response to this concern, researchers have emphasized the importance of responsible software engineering~\cite{schieferdecker2020responsible,bennaceur2024responsible} and the need to operationalize human-centered values throughout the software development lifecycle, from requirements specification to software testing~\cite{shahin2022operationalizing,mougouei2018operationalizing}.
Recently, the literature has highlighted the need for architectural pattern and design techniques to operationalize the principles of responsible AI and align software to human-centered values~\cite{lu2022roadmap}.

A common approach to align agentic AI systems is to treat value guarantees as post-hoc constraints by introducing system-level guardrails, by filtering the final output or enforcing policies through human-in-the-loop intervention applied to an otherwise unconstrained architecture~\cite{kim2026attack,shamsujjoha2025swiss}.
However, these mechanisms depend on safeguards defined \textit{a priori} and inspect outputs only after upstream transformations, when biased or harmful information have already been introduced, aggregated, and propagated.
This limitation suggests that value preservation cannot be delegated to post-hoc validation only:
architectural decisions such as coordination mechanisms, communication protocols, and system topologies play a central role in determining how values are ultimately realized, complementing requirements specification and system testing.
Therefore, as system behavior emerges from the interactions between autonomous agents, the design of MAS architectures becomes central, and the following research question emerges: \textit{how can architectural design be leveraged to support human-centered values in the era of agentic AI?}

This vision paper highlights the importance of MAS architectures to satisfy privacy, fairness, and pluralism in a structured way, by addressing recurring human-centered requirements in the form of reusable value-preserving architectural patterns.
While previous works have proposed general agentic AI pattern catalogs~\cite{liu2025agent} and studied the need for security~\cite{raza2026trism}, privacy~\cite{li2025123,yagoubi2026agentleak}, fairness~\cite{borah2024implicit,aird2024dynamic}, and pluralism~\cite{ashkinaze2025plurals,feng2024modular} in the context of MAS, this paper investigates the interplay between agentic architectural choices and human-centered values, representing a first step towards responsible software engineering for the design of MAS.
We see this as a concrete step for engineers to retain control over human values in the systems they build.
This paper makes the following contributions:
\begin{itemize}
  \item[\textbf{1.}] We frame the \textbf{principal challenges of responsible software engineering} in the context of MAS, highlighting the interplay between human-centered values in agentic AI systems and architectural design.
  \item[\textbf{2.}] We discuss three \textbf{value-preserving architectural patterns} -- Federated Silos Coordination, Peer-to-Peer Deliberation, and Plan-Triggered Guard Agents, preserving privacy, pluralism, and fairness respectively -- presenting their design, agent responsibilities, and coordination flow.
  \item[\textbf{3.}] We illustrate the architectural patterns with \textbf{concrete use-case scenarios}, showing how their design structurally supports the target value.
\end{itemize}

\noindent The rest of the paper is organized as follows.
Section~\ref{sec:background} provides an overview of agentic AI and related work in the context of responsible software engineering.
Section~\ref{sec:architectures} describes the three value-preserving architectural patterns for privacy, pluralism, and fairness. 
Section~\ref{sec:use-cases} illustrates the proposed design on real-world scenarios.
Finally, Section~\ref{sec:conclusions} concludes the paper by outlining research directions towards the creation of a unified catalog of value-preserving architectural patterns for MAS.

\section{Background} 
\label{sec:background}
%
\paragraph{Agentic AI and MAS}
An agentic AI system is an LLM-powered entity that (i)~acts autonomously without step-by-step human instruction, (ii)~plans by decomposing a high-level goal into ordered sub-tasks, and (iii)~uses tools -- APIs, databases, code interpreters -- to execute them.
Cognitive-architecture treatments map the memory, action, and decision modules any language agent requires, clarifying why planning and tool use are constitutive features of agency rather than add-ons~\cite{sumers2024cognitive}; complementary work shows that iterative self-reflection over execution traces, rather than a single forward pass, is what let agents recover from failures on long-horizon tasks~\cite{shinn2023reflexion}.
Already at the single-agent level this autonomy trades transparency for capability: reasoning unfolds across many internal steps and tool calls that do not surface to the human, thus reducing the interpretability of the rationale behind a given action.
 
MAS amplify this opacity along three axes. First, \textit{behavioral} opacity: coordinating several specialized agents under a shared objective yields decomposition, per-domain skill, and resilience to localized failures~\cite{hong2023metagpt}, but the resulting behavior is an emergent property of inter-agent interaction rather than of any inspectable component, so it cannot be read off from any single agent.
Second, \textit{attributional} opacity: when an outcome is produced jointly by many autonomous agents exchanging intermediate results, responsibility for it -- an exposed record, a discarded viewpoint, a biased inference -- is diluted across the network and hard to trace back to a single locus~\cite{chan2023harms,li2025123}.
Third, \textit{control} opacity: in the canonical orchestrator topology every sub-task is mediated by natural-language instructions flowing from one node, so a semantic failure there propagates silently to the whole network whenever no human is
observing~\cite{greshake2023not}, and the depth of the interaction graph leaves few points at which a human could meaningfully intervene.
The net effect is that robustness to faults does not transfer to robustness against misaligned objectives, and the engineer's ability to \textit{observe, attribute, and arrest} undesired behavior erodes precisely as capability grows -- strategies.
Opacity is thus the structural liability that makes human values preservation so difficult to guarantee by external means, a gap only partially closed by human-in-the-loop~\cite{mosqueira-rey2023humanintheloop}. This motivates moving the guarantee into the architecture itself.

\paragraph{Responsible AI and Software Engineering}
The growing societal impact of software systems, and AI-based software in particular, has shifted software engineering research and practice towards explicitly incorporating ethical considerations throughout the development lifecycle.
\textit{Ethics-aware software engineering}~\cite{aydemir2018roadmap} proposes to systematically capture and analyze stakeholder values during software specification and development, including trustworthy principles such as fairness, interpretability, and safety.
Building on this direction, a bridge between software engineering and responsible AI has been proposed~\cite{lu2022roadmap,zhu2023software} to guide the design, development, and deployment of AI-based software while accounting for their broader societal impact and ensuring alignment with human-centered values.
\textit{Responsible software engineering} elevates the centrality of human-centered values such as privacy, fairness, and explainability to first-class requirements and design goals, treating them with the same rigorous methodology traditionally reserved for functional requirements~\cite{bennaceur2024responsible,schieferdecker2020responsible}.
This perspective resonates with other initiatives such as value sensitive design~\cite{friedman2017survey,zuber2024valuesensitive}, which advocate for a principled and systematic account of stakeholder values throughout the design process.
Operationalizing human values in software systems denotes the process of identifying relevant user values and translating them into accessible, concrete concepts that can be implemented, validated, verified, and measured within software systems~\cite{shahin2022operationalizing,whittle2021case,mougouei2018operationalizing,ferrario2023applying}.
More recently, the emergence of LLM-based software agents has given rise to the notion of \textit{socio-critical systems}, a new category of software systems whose behavior can have significant ethical and societal consequences, underscoring the need to embed human-centered values not only at runtime, but already during their development stages~\cite{baresi2024conceptual}.

\section{Value-Preserving Architectures}
\label{sec:architectures}
We present three architectural patterns, each targeting a specific human-centered value: Federated Silos Coordination for privacy (Section~\ref{sec:privacy}), Peer-to-Peer Deliberation for pluralism (Section~\ref{sec:pluralism}), and Plan-Triggered Guard Agents for fairness (Section~\ref{sec:fairness}).
For each pattern, we first define the target value it aims to preserve and give an overview of the corresponding architecture. We then specify the pattern in a context--problem--solution form~\cite{bass2003software,camilli2025software}.
The \textit{context} describes a situation that gives rise to a problem, the \textit{problem} outlines the issues and key constraints that arise in that context, and the \textit{solution} describes the architecture that address it.

\subsection{Federated Silos Coordination Pattern}
\label{sec:privacy}

 \begin{figure}[tbp]
    \centering
    \includegraphics[width=0.99\linewidth]{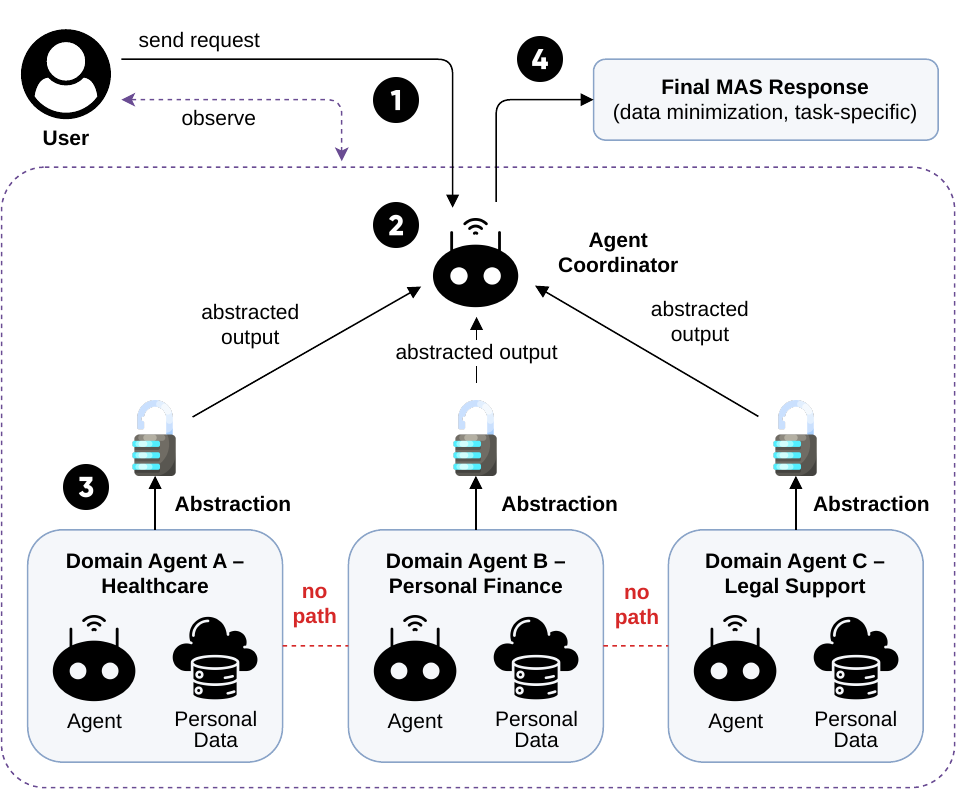}
    \caption{Federated Silos Coordination Pattern}
    \Description{A diagram describing a multi-agent system architecture to support privacy.}
    \label{fig:privacy}
\end{figure}

Privacy guarantees the user's control over the access and use of their
personal data.
In the context of MAS, we extend this traditional notion to include fine-grained control over what is shared with each autonomous agent: personal data must be governed not only at the boundary between the system and the outside world, but also within the system.
Figure~\ref{fig:privacy} presents the \textit{Federated Silos Coordination} architectural pattern, in which a coordinator agent orchestrates the overall workflow by collecting and combining only the minimal output required from each domain agent, without accessing or storing the underlying data itself.
Domain agents never exchange information among themselves: the only information leaving a silo consists of minimal, task-specific abstractions, in accordance with the principle of \textit{data minimization}, which requires using only the information strictly necessary for the task at hand.

\begin{discussionbox}{Federated Silos Coordination Pattern}
\textbf{Context.} A MAS operating over disjoint domains of sensitive, subject-specific information, and a user-facing task
that may require combining cross-domain information.

\textbf{Problem.} Requests must be
routed only to the minimal subset of domain agents required for completion. When cross-domain combination is needed, any direct exchange between domain agents risks leaking sensitive data beyond
the domain to which it belongs. Filtering only the final answer is
insufficient: once a detailed value crosses a domain boundary, it can
influence subsequent intermediate reasoning, making its effects
impossible for a final-output filter to reliably detect or reverse.
The architecture must therefore guarantee that (i)~no domain agent
ever observes another domain's records, (ii)~any value leaving a
domain carries no more precision than the task requires, and
(iii)~data minimization is promoted while retaining the information necessary to complete the task.

\textbf{Solution.} Organize the system around a single
\textit{coordinator} and a set of \textit{domain agents}, providing an architectural enforcement point for privacy. First, at
each domain agent, interpose a local \textit{abstraction step}: every value the agent is about to externalize
is generalized before emission (e.g., a scalar is bucketed into a
range, a detailed attribute is mapped onto a coarser category).
Second, restrict the communication topology to a hub-and-spoke form:
abstracted representations are emitted only to the coordinator and
never to sibling domain agents, so no direct inter-domain channel
exists. The coordinator routes each request only to the minimal
subset of specialized agents the task requires and integrates their
intermediate outputs into a coherent response; this integration is
\textit{selective}, governed by data minimization: only the signals strictly necessary for task completion are combined, and
subject-specific details not required by the final output are never exposed.
\end{discussionbox}

\subsection{Peer-to-Peer Deliberation Pattern}
\label{sec:pluralism}

\begin{figure}[tbp]
    \centering
    \includegraphics[width=0.9\linewidth]{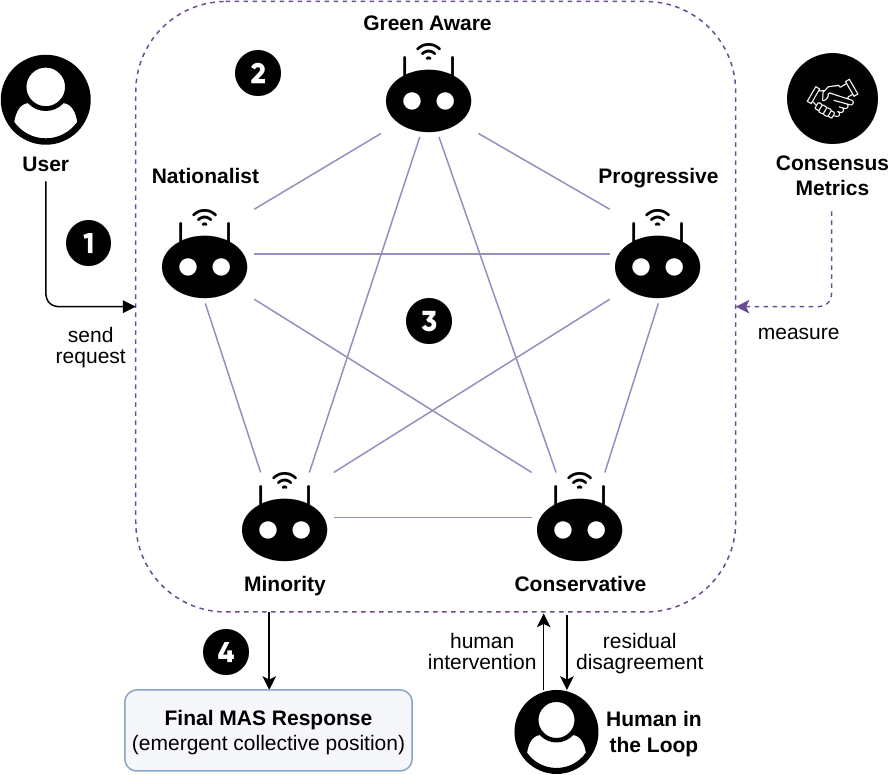}
    \caption{Peer-to-Peer Deliberation Pattern}
    \Description{A diagram describing a multi-agent system architecture to preserve pluralism and diversity.}
    \label{fig:pluralism}
\end{figure}

Pluralism is the belief that diverse groups are represented equally and fairly within the system, promoting diversity and active cooperation.
Figure~\ref{fig:pluralism} presents an overview of the \textit{Peer-to-Peer Deliberation} architectural pattern where pluralism each agent communicates directly with each other, no central orchestrator entity controls the deliberation, and the collective position
emerges from iterative exchange rather than from aggregation by an authoritative node.
Each agent impersonates a distinct deliberative position and no node holds final authority over the output.

\begin{discussionbox}{Peer-to-Peer Deliberation Pattern}
\textbf{Context.} A MAS in which a user-facing task requires
considering multiple legitimate perspectives, values, or viewpoints,
and no single agent's opinion should be given overriding authority.

\textbf{Problem.} Any centralized coordination point risks
privileging one perspective over the others. At the same time, a
fully decentralized deliberation among opinionated agents risks
never converging, with each agent defending its own position
indefinitely. The architecture must therefore guarantee that
(i)~no single agent can determine the outcome, (ii)~diverse
deliberative positions are structurally represented, and (iii)~the
deliberation terminates, either in a shared position or in an
explicit account of the residual disagreement.

\textbf{Solution.} Remove any privileged coordinator and arrange
role-bearing agents in a fully decentralized, complete communication
graph: every agent communicates directly with every other. Assign
each agent a distinct deliberative position, so that the space of
perspectives is structurally covered. To resolve the tension between
pluralism and convergence, define a \textit{consensus metric} that
quantifies how far the agents currently are from a shared position.
The metric serves two purposes: it provides a measurable stopping
condition, turning an open-ended exchange into a process with
readable progress; and it enables intervention in the agents'
feedback loops: when the metric signals stalling or polarization,
the system nudges agents towards compromise without imposing any
single opinion. The objective is not to manufacture agreement at any
cost, which would itself erode pluralism, but to reach well-founded
compromise whenever possible. When no such compromise is reached,
the system does not force convergence: the residual disagreement,
together with the positions held by each agent, is surfaced to the
human user, who retains final decision authority.
\end{discussionbox}

\subsection{Plan-Triggered Guard Agents Pattern}
\label{sec:fairness}

 \begin{figure}[tbp]
    \centering
    \includegraphics[width=0.99\linewidth]{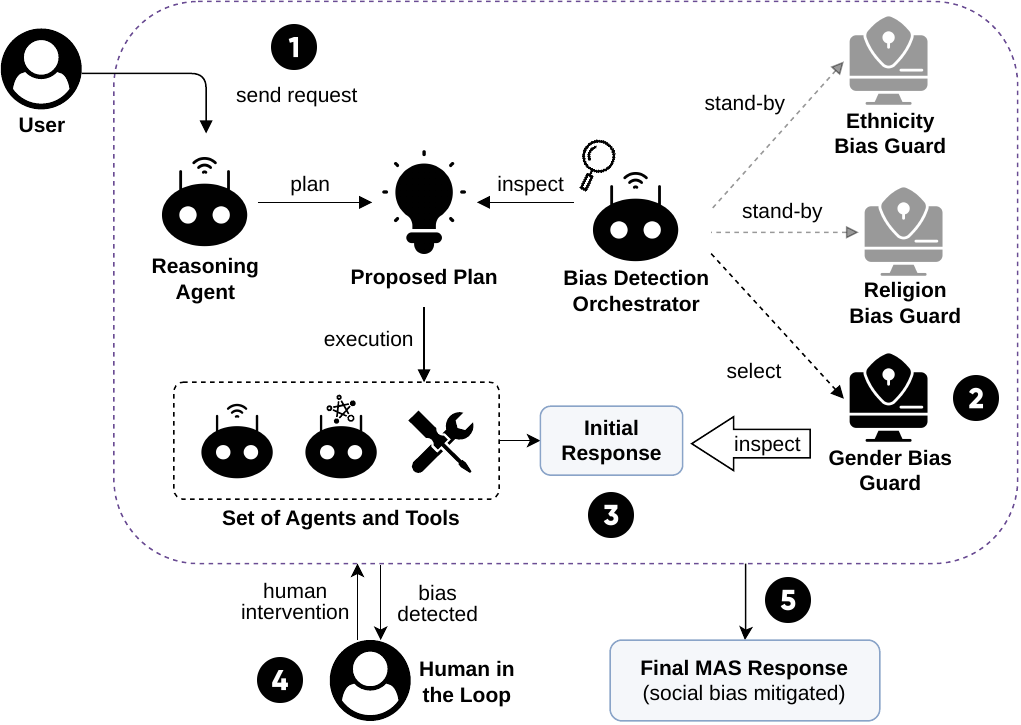}
    \caption{Plan-Triggered Guard Agents Pattern}
    \Description{A diagram describing a multi-agent system architecture to preserve fairness.}
    \label{fig:fairness}
\end{figure}

Fairness requires that no group or individual subject is systematically disadvantaged or discriminated by the system.
Figure~\ref{fig:fairness} illustrates the \textit{Plan-Triggered Guard Agents} pattern where fairness is promoted by interposing specialized guard agents into the processing flow, each agent targeting a specific bias type (e.g., gender, ethnicity, religion).
Rather than inspecting the final output where bias is hard to detect, guards intercept and correct bias at runtime, before it contaminates the downstream steps. Guards are modular nodes, attached to but separable from the core pipeline.

\begin{discussionbox}{Plan-Triggered Guard Agents Pattern}
\textbf{Context.} A MAS supporting decision-making tasks that affect
individuals or groups (e.g., hiring, credit, healthcare), where the
system output is susceptible to one or more classes of bias.

\textbf{Problem.} Bias can emerge at any phase of the
decision-making process, and even a single biased step may
compromise the fairness of the entire outcome: a biased intermediate
action propagates through subsequent agentic steps and shapes the
result in ways that a post-hoc check might not be able to reverse.
At the same time, not every task carries the same risk: applying every possible bias check to every step is impractical. The architecture must therefore guarantee that (i)~bias detection is targeted to the specific risks of the decision-making process rather than applied indiscriminately, (ii)~each detected bias is handled by a component specialized for that bias class to enhance modularity, (iii)~mitigation occurs before performing any action (e.g. tool invocation, answer generation, skill selection).

\textbf{Solution.} Separate the pipeline into three phases:
\textit{plan}, \textit{review}, and \textit{execute}. In the plan phase,
the agent reasons over the task and emits a proposed action plan without carrying out any action. In the review phase, a
\textit{bias-detection orchestrator} inspects the plan and selectively
activates specialized \textit{guard agents}, each designed to detect
and mitigate a specific class of bias; only the guards warranted by
the structure and risk profile of the plan are employed. Each
activated guard verifies the presence or absence of its target bias and intervenes before any action; when a detected bias cannot be automatically mitigated, the
plan is flagged and escalated to a human reviewer, who retains final decision authority. Because the guard agents are decoupled from the
core pipeline, they can be configured per domain, updated, or replaced independently.
\end{discussionbox}

\section{Patterns in Practice: Application Examples}
\label{sec:use-cases}
In this section, we assess the architectural patterns from the previous section on different use-case scenarios, showing how their design structurally sustain the target values.
\subsection{Federated Silos Coordination for Privacy}
Consider a general-purpose MAS designed to support personalized decision-making across multiple domains, e.g., healthcare and financial planning, by processing user personal data.
The primary stakeholder is the end user, entitled to receive decision-support and personalized suggestions with privacy preservation and guarantees of data minimization.
In conventional MAS architectures, agents may share or access unnecessary cross-domain personal data, maximizing the use of data.
However, in a personalized healthcare scenario, a medical agent should not access legal or financial personal data: previous legal disputes related to hospital bills, outstanding debts, or low income should not influence medical recommendations, which must rely exclusively on health-related information.

\textit{Scenario Workflow.}
The user requests personalized decision support for a suitable treatment for hypertension (step~\ding{202} in Figure~\ref{fig:privacy}).
Upon receiving the request, the agent coordinator identifies the medical domain and dispatches the task to the medical agent (step~\ding{203}), which accesses only the user's clinically relevant information (e.g., medical history, current medications, and diagnostic results) in accordance to the principle of data minimization.
The agent invokes external tools to generate treatment recommendations, e.g., machine-learning models for precision medicine (step~\ding{204}), without accessing personal information about unrelated domains, e.g., financial status or legal records.
The final recommendation is returned to the coordinator, which outputs the result to the user (step~\ding{205}).
Suppose the user explicitly requests a treatment recommendation considering financial affordability: the coordinator would independently dispatch the request to both the medical and financial agents.
Each agent processes only the data required for its domain and produces independent, abstracted responses on available treatments and personal finance.
The coordinator presents the outcomes to the user without exchanging or merging personal data across domains, preventing information leakage by design.

\subsection{Peer-to-Peer Deliberation for Pluralism}
Consider a MAS designed to retrieve, aggregate, and summarize online news on current events.
The main stakeholders include the end users, who consume the generated summaries, and the news providers, whose diverse perspectives should be equally represented.
Diversity is the key system requirement to ensure that generated summaries reflect a plurality of viewpoints rather than only dominant or polarized narratives.
In conventional agentic architectures, information retrieval and summarization are typically centralized.
As a result, the system may over-represent majority viewpoints, either because they are more prevalent in the retrieved sources or due to the central LLM tendency to produce summaries aligned with its internal preferences, potentially overlooking minority or dissenting perspectives.

\textit{Scenario Workflow.}
A user requests a summary of the latest international climate summit (step~\ding{202} in Figure~\ref{fig:pluralism}).
The user request is forwarded to multiple peer news-retrieval agents.
Each agent independently gathers information from a distinct set of news sources, producing a preliminary summary that reflects their alignment:
the nationalist agent emphasizes the need for energy autonomy at the expense of climate impact, the green-aware agent stresses the need for climate action, and the minority agent highlights concerns from developing countries (step~\ding{203}).
The agents exchange and discuss their findings through a deliberation phase, allowing alternative, and minority viewpoints to be surfaced rather than suppressed by a single centralized summarization process (step~\ding{204}).
The resulting perspectives represent both majority opinions and minority opinions promoting representation of diverse viewpoints and reduces the risk of over-emphasizing dominant narratives (step~\ding{205}).

\subsection{Plan-Triggered Guard Agents for Fairness}
Consider an agentic decision-support system for human resource management.
The system supports multiple tasks through external tool invocation, e.g., automated resume parsing, candidate evaluation, or generating reports about employees.
While fairness and equal treatment are critical for bias-sensitive tasks such as resume parsing~\cite{dolci2023improving,bolukbasi2016man}, it is less relevant for descriptive tasks, motivating selective activation of fairness-preserving mechanisms.
In this context, the stakeholders include job applicants, whose opportunities may be affected by automated decisions, and the company, whose reputation and regulatory compliance depend on fair hiring practices.
In conventional architectures, LLM-based agents may inherit biases from historical training data and reproduce discriminatory patterns through statistical associations.
Furthermore, while agentic workflows typically include system-level guardrails and external detection systems, biases introduced by one agent may propagate or even be amplified across reasoning stages within the system.

\textit{Scenario Workflow.}
A recruiter requests to rank candidates for a software engineering position (step~\ding{202} in Figure~\ref{fig:fairness}).
Two candidates are considered: Candidate A is male, while Candidate B is female.
Upon identifying the task as fairness-critical, the bias-detection orchestrator activates the gender guard agents, which inspect the intermediate ranking and identify that sensitive attributes may influence the recommendation (step~\ding{203}).
Although Candidate B has prior professional experience that better match the job requirements, the agentic workflow initially ranks Candidate A higher, reflecting the historical bias of software engineering been predominantly associated with males (step~\ding{204}).
The workflow is therefore flagged and the task requires the intervention of the human recruiter (step~\ding{205}), who reviews the candidates and ultimately selects Candidate B, preventing unfair outcomes (step~\ding{206}).

\section{Conclusions and Research Directions}
\label{sec:conclusions}
In this paper, we investigated how architectural choices in MAS can reinforce human-centered values by design.
We introduced an initial set of MAS architectural design patterns to support privacy, pluralism, and fairness, laying the foundation for a unified catalog of patterns to develop human-centered MAS.
Future research will investigate how different agentic patterns can be composed to satisfy multiple value requirements simultaneously, examining the synergies, trade-offs, and limitations that arise when coordinating architectures designed to optimize different values.
Moreover, we plan to address the problem of adapting agentic systems to evolving stakeholder requirements and the potential shift of values at system runtime~\cite{gavidia-calderon2022what,bennaceur2023valuesruntime}.
Finally, the definition of metrics is another crucial challenge for evaluating value preservation, representing an important step towards operationalizing values in software systems~\cite{shahin2022operationalizing}, e.g., metrics to measure pluralistic alignment~\cite{sorensen2024position}.

\begin{acks}
This work is supported by ARMADA, funded by the European Union's Horizon Europe  Marie Sk\l{}odowska-Curie Actions (MSCA) under grant No.\ 101168951, by the Vienna Science and Technology Fund (WWTF) grant Nos.\ 10.47379/VRG18013, 10.47379/ICT25032, 10.47379/NXT22018, 10.47379/ICT2201, 10.47379/DCDH001, and by the Austrian Science Fund (FWF) grant No.\ 10.55776/COE12.
\end{acks}


\bibliography{references}

\end{document}